\documentclass{article}

\usepackage{microtype}
\usepackage{graphicx}
\usepackage{subcaption}
\usepackage{booktabs} % for professional tables

\usepackage{hyperref}
\usepackage[inkscapelatex=false]{svg}

\usepackage[accepted]{icml2024}

\usepackage{amsmath}
\usepackage{amssymb}
\usepackage{mathtools}
\usepackage{amsthm}

\usepackage[capitalize,noabbrev]{cleveref}

\theoremstyle{plain}

\theoremstyle{definition}

\theoremstyle{remark}

\usepackage[textsize=tiny]{todonotes}

\icmltitlerunning{Snail classification with multimodal triplet networks}
\defcitealias{iucn_international_union_for_conservation_of_nature_habitats_2012}{IUCN, 2012}

\begin{document}

\twocolumn[
\icmltitle{Classification of freshwater snails of the genus \emph{Radomaniola} with multimodal triplet networks}



\icmlsetsymbol{equal}{*}

\begin{icmlauthorlist}
\icmlauthor{Dennis Vetter}{CVAI}
\icmlauthor{Muhammad Ahsan}{CVAI}
\icmlauthor{Diana Delicado}{SBL}
\icmlauthor{Thomas A. Neubauer}{SNSB}
\icmlauthor{Thomas Wilke}{SBL}
\icmlauthor{Gemma Roig}{CVAI}
\end{icmlauthorlist}

\icmlaffiliation{CVAI}{Computational Vision and Artificial Intelligence Lab, Goethe University Frankfurt, Frankfurt am Main, Germany}
\icmlaffiliation{SBL}{Department of Animal Ecology \& Systematics, Justus Liebig University, Giessen, Germany}
\icmlaffiliation{SNSB}{Bavarian State Collection for Palaeontology and Geology, SNSB, Munich, Germany}

\icmlcorrespondingauthor{Dennis Vetter}{vetter@em.uni-frankfurt.de}

\icmlkeywords{applied machine learning, biology, taxonomy, triplet network}

\vskip 0.3in
]



\printAffiliationsAndNotice{}  % leave blank if no need to mention equal contribution

\begin{abstract}
In this paper, we present our first proposal of a machine learning system for the classification of freshwater snails of the genus \emph{Radomaniola}. We elaborate on the specific challenges encountered during system design, and how we tackled them; namely a small, very imbalanced dataset with a high number of classes and high visual similarity between classes. We then show how we employed triplet networks and the multiple input modalities of images, measurements, and genetic information to overcome these challenges and reach a performance comparable to that of a trained domain expert.
\end{abstract}

\section{Introduction}
\label{sec:introduction}
\begin{figure}[ht]
    \centering
    \begin{subfigure}{0.45\linewidth}
        \centering
        \includegraphics[width=0.95\linewidth]{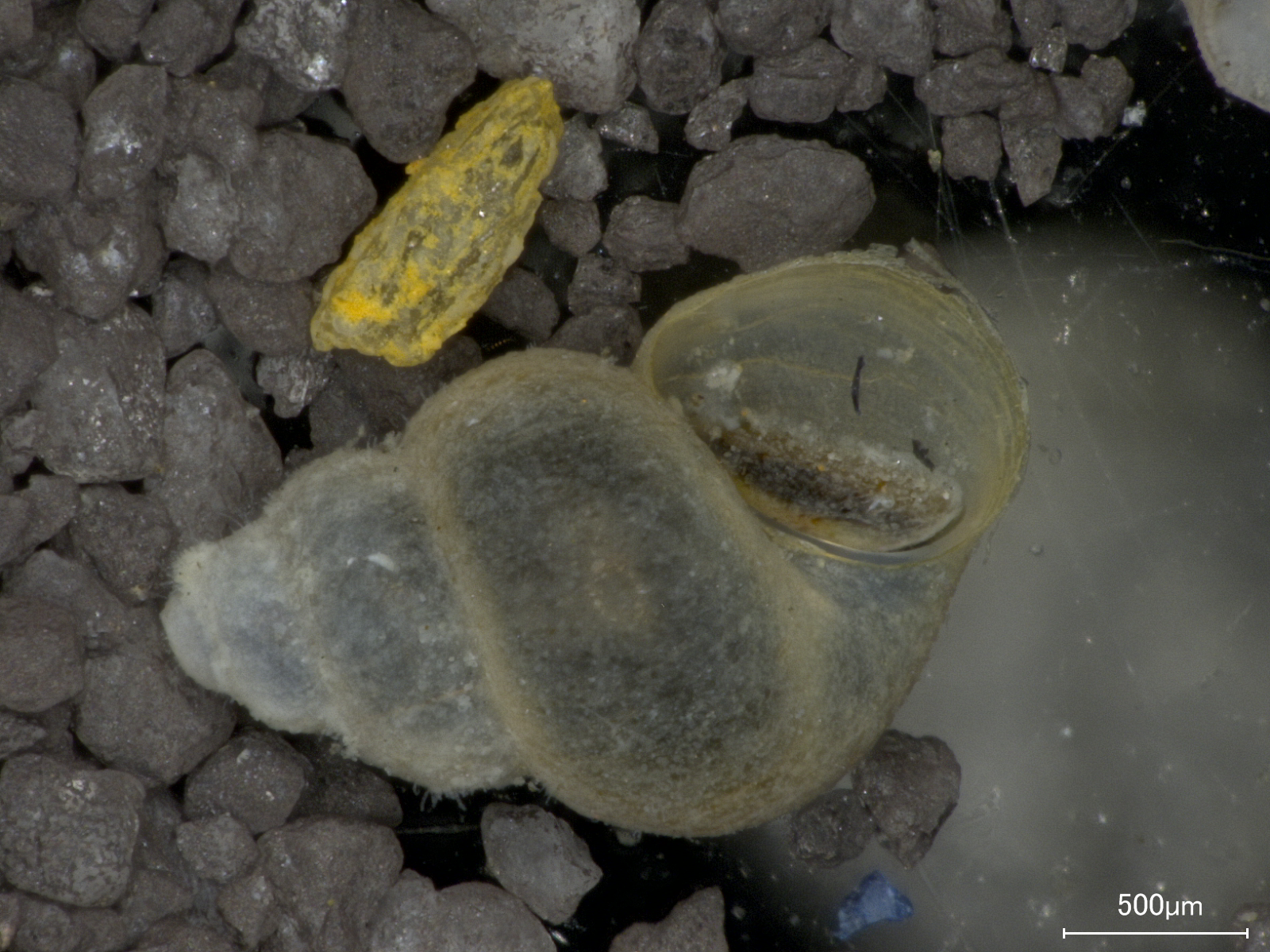}
    \end{subfigure}
    \begin{subfigure}{0.45\linewidth}
        \centering
        \includegraphics[width=0.95\linewidth]{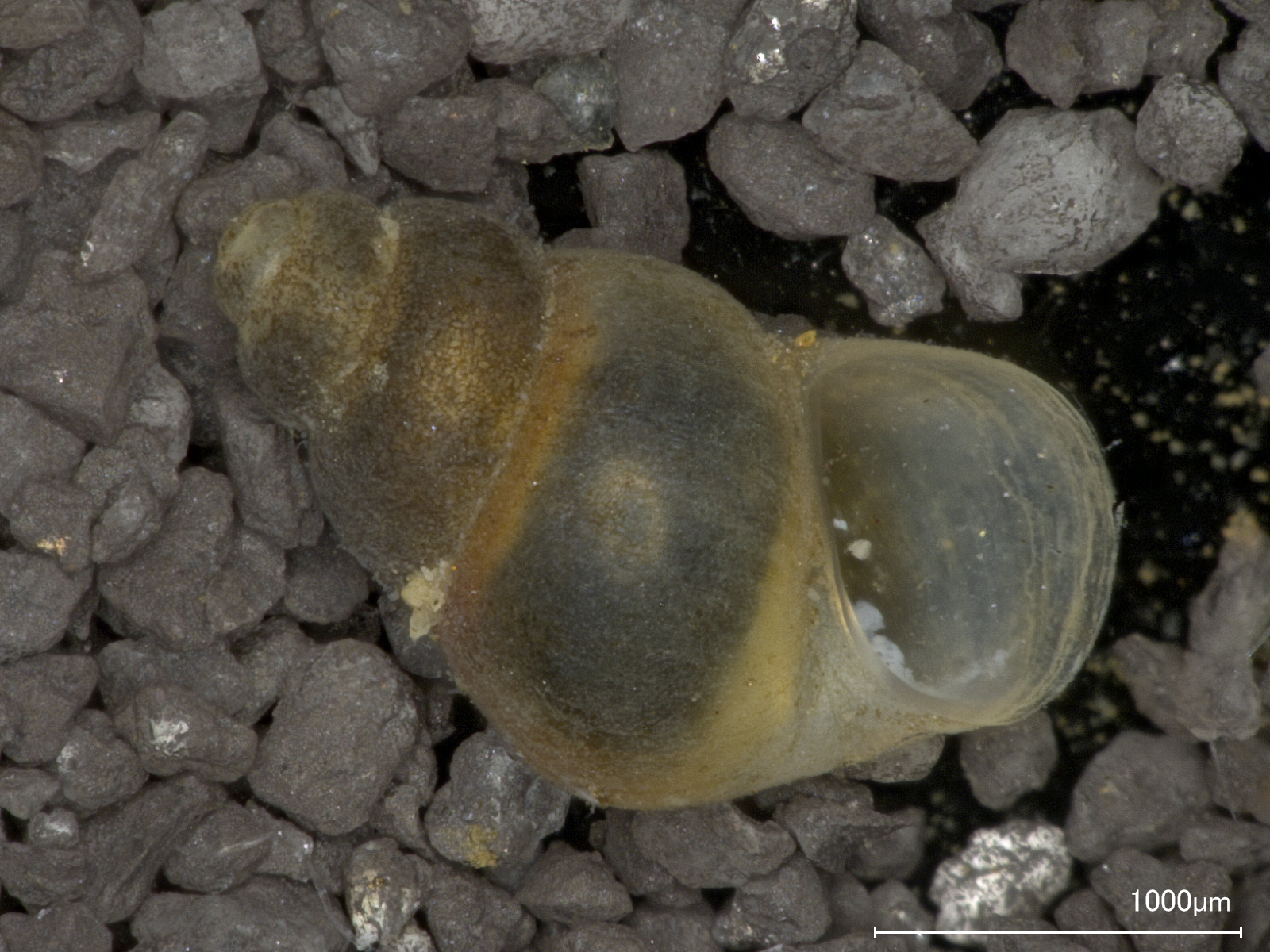}
    \end{subfigure}
    
    \begin{subfigure}{0.45\linewidth}
        \centering
        \includegraphics[width=0.95\linewidth]{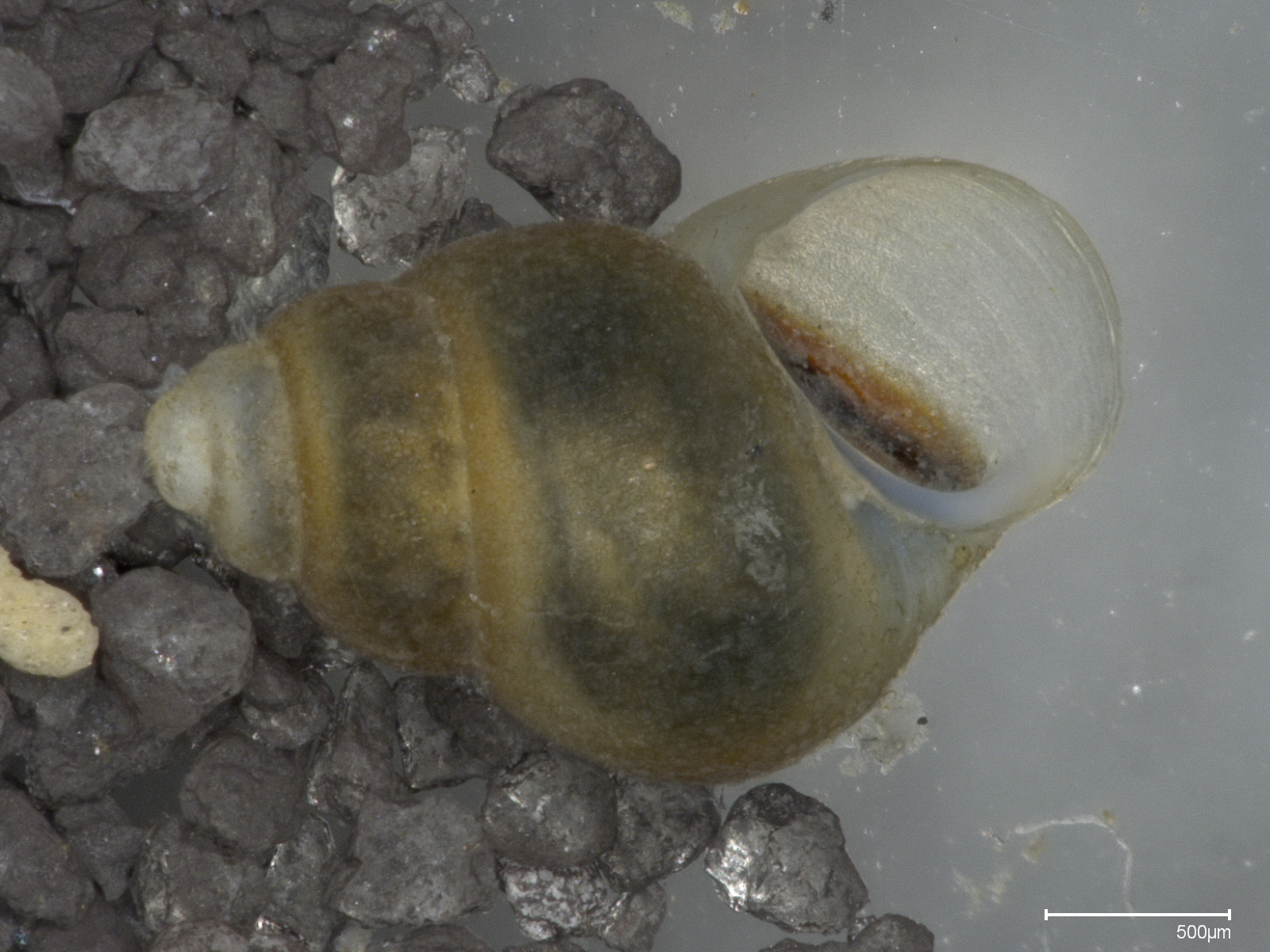}
    \end{subfigure}
    \begin{subfigure}{0.45\linewidth}
        \centering
        \includegraphics[width=0.95\linewidth]{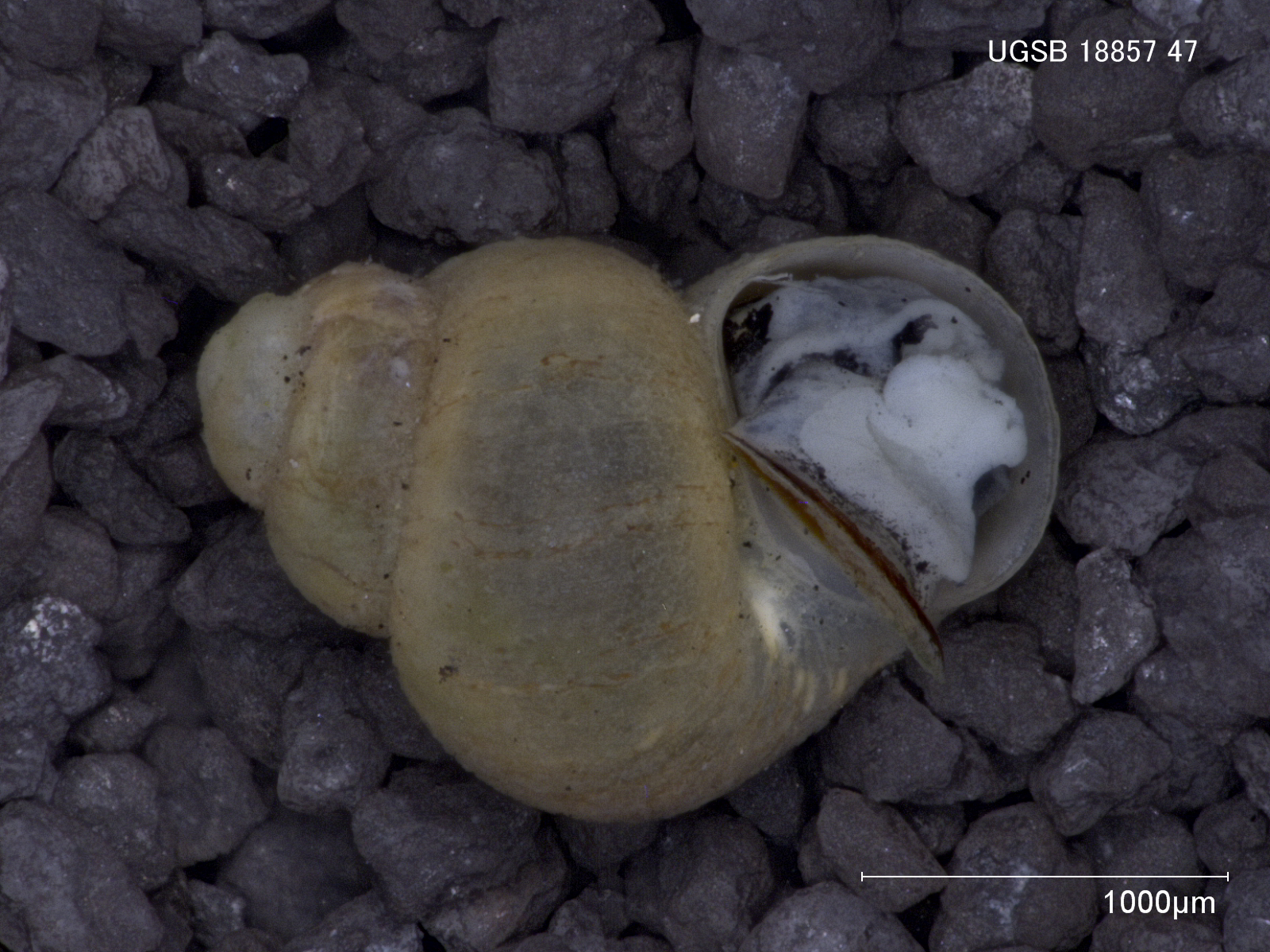}
    \end{subfigure}
    
    \begin{subfigure}{0.45\linewidth}
        \centering
        \includegraphics[width=0.95\linewidth]{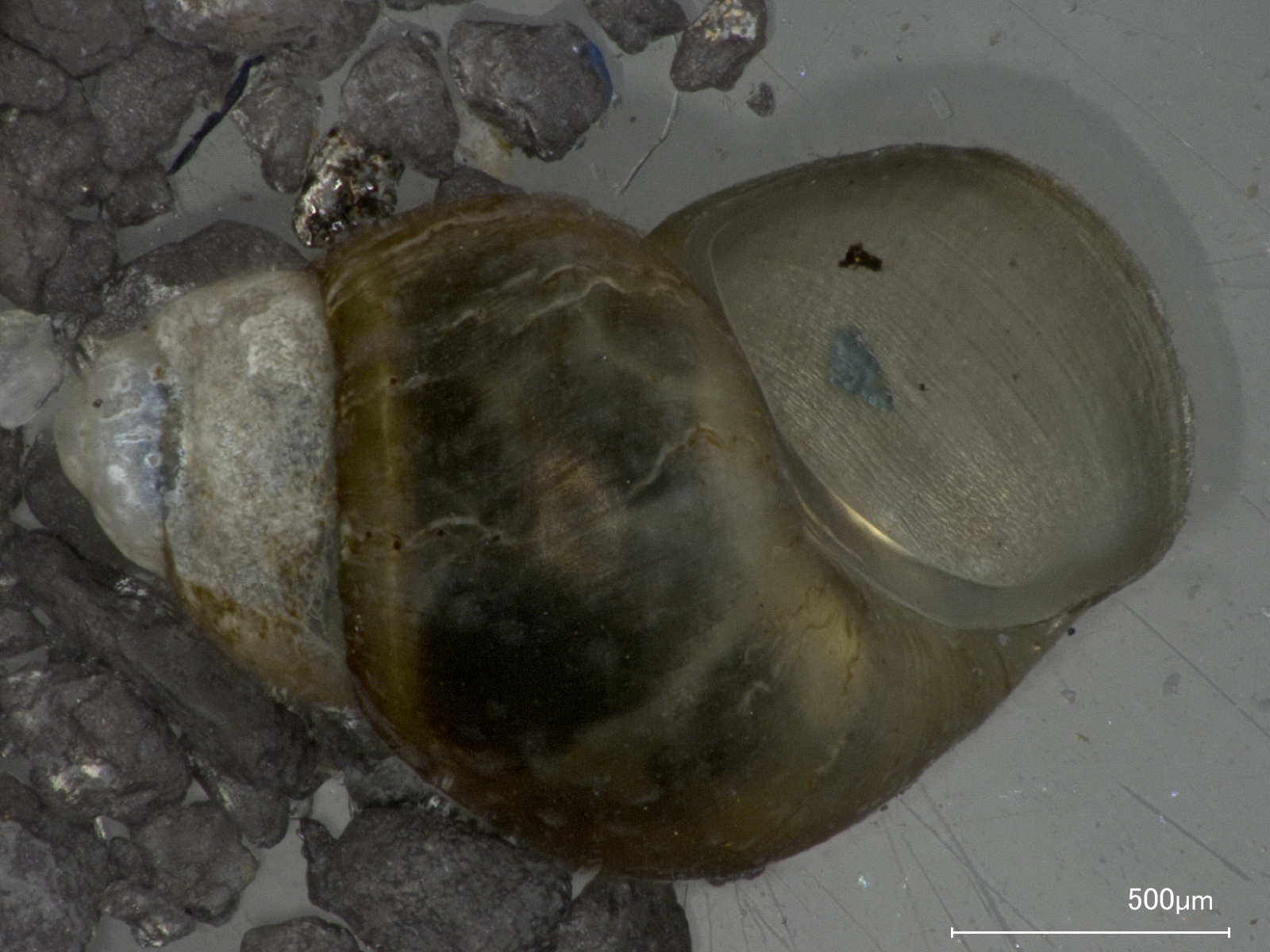}
    \end{subfigure}
    \begin{subfigure}{0.45\linewidth}
        \centering
        \includegraphics[width=0.95\linewidth]{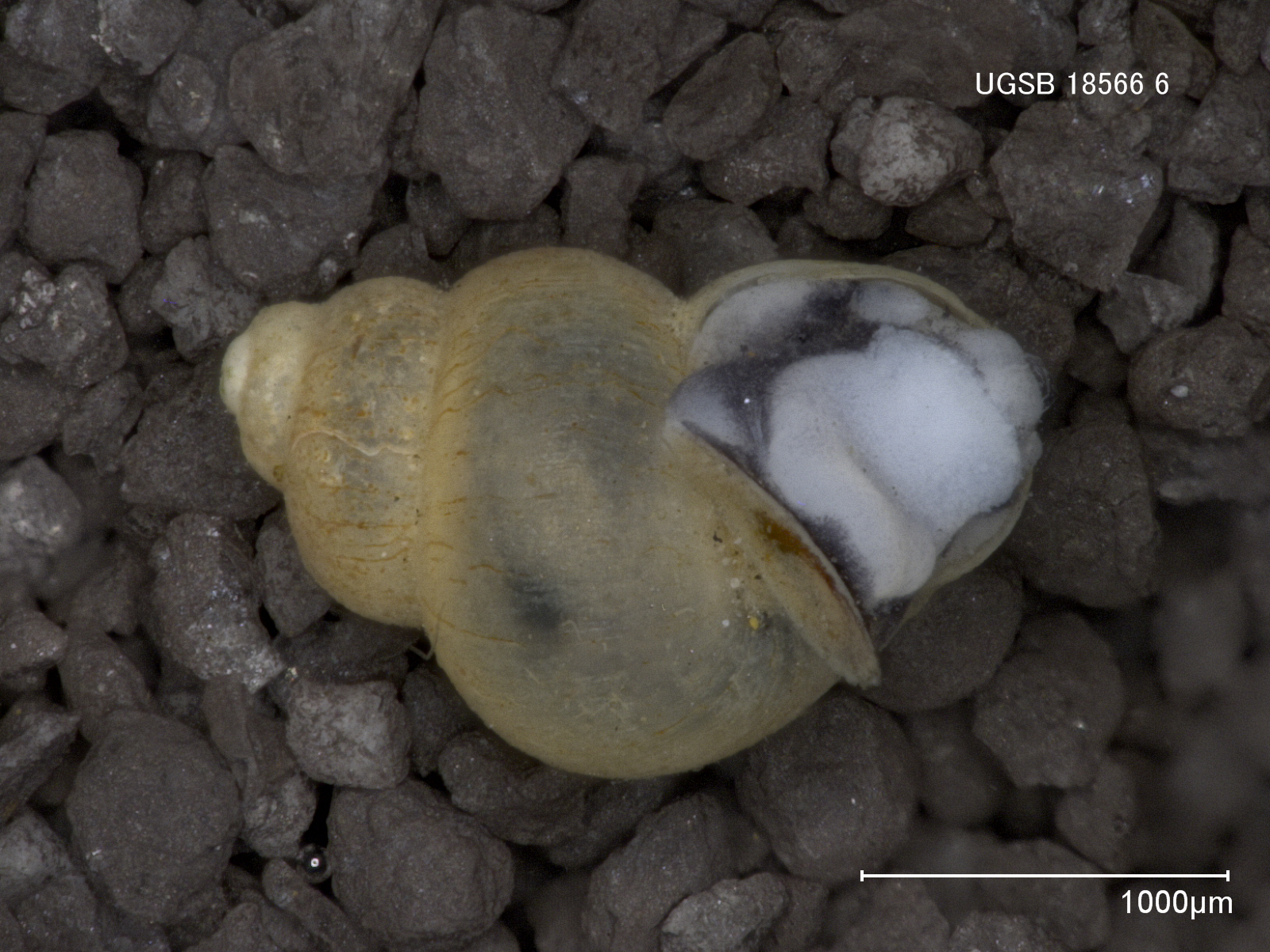}
    \end{subfigure}
    \caption{Example specimens from six different \emph{Radomaniola} species. Left: \emph{R. curta}, \emph{R. mostarensis}, \emph{R. seminula}. Right: \emph{R. jovanovskae}, \emph{R. nachtigallae}, \emph{R. szarowskae}. To an untrained observer the species appear very similar, differing only in minute details.}
    \label{fig:snails}
\end{figure}

\emph{Radomaniola} is a genus of gastropods (`snails') that live in springs and other flowing waters in the Balkan region \cite{boeters_radomaniolagrossuana_2017}. 
Due to their minute size (2-4 mm long) and featureless shells, the taxonomy (species classification) of these snails today is primarily based on genetic and anatomical data \cite{delicado_shell_2022}. 
As illustrated in \cref{fig:snails}, the differences in shell morphology between species are only minor, particularly to the untrained eye. 
Consequently, classifying \textit{Radomaniola} specimens is complex and time-consuming, as the subtle morphological differences are challenging to detect, even for human experts. 
Traditional methods involve multiple, labor-intensive steps, including transferring specimens to a laboratory, detailed examination under a binocular microscope, specimen dissection, and meticulous comparison with other species within the genus. 
Additionally, the limited number of experts and their lengthy training process can further delay species identification.

In the following, we demonstrate how we employed multimodal triplet networks to develop a system that can learn from images, measurements and genetic information, while at the same time working around a small and heavily imbalanced dataset to reach expert-level classification accuracy. 
This classification module is part of a larger work-in-progress system that we plan to use in the future to support taxonomists in their work by providing them with a readily available objective, AI-based assessment, and in making their knowledge more accessible.

\section{Related Work}
Our currently best-performing system leverages multiple input modalities, specifically images, measurements, and genetic information. We use these inputs to simultaneously learn an intermediate representation that captures similarities between inputs, and how to use this intermediate representation for the final classification task. In the following we will therefore briefly cover the fundamentals of multimodal learning, similarity learning, and multi-task learning.

\subsection{Multimodal Learning}
In multimodal learning, the system learns from multiple input modalities simultaneously. Instead of relying only on tabular data, text, or images, the system is integrating and learning from a combination of these modalities. The rationale is that each modality captures different aspects of the data, therefore learning from multiple modalities can potentially lead to a more in-depth understanding \cite{baltrusaitis_multimodal_2017}. A key challenge in multimodal deep learning is how to efficiently represent the different modalities. A common solution involves creating a separate set of layers for each modality to compute intermediate representations, which are then projected into a joint space. This joint representation is then used for a downstream task, such as classification \cite{gao_survey_2020, baltrusaitis_multimodal_2017}. Previous research has shown that this approach allows for the entire network to be trained end-to-end to simultaneously learn an efficient representation of the different modalities and performing the task at hand \cite{ngiam_multimodal_2011, mroueh_deep_2015, ouyang_multi-source_2014, nguyen_multimodal_2019, hong_multimodal_2015}.
In our work, we have different modalities of data available: images, measurements and genetic information. According to the domain experts these modalities complement each other and the decision is usually made taking multiple into account and not one alone when more are available.

\subsection{Similarity Learning}
While deep learning systems can be incredibly powerful, they typically require large amounts of data to perform well. However, in many practical applications, acquiring such large datasets demands considerable effort and time from domain experts, which is often not feasible. In our case this would require the domain experts to either travel to the remote locations where specimens were previously collected, or to identify new locations where \emph{Radomaniola} snails occur; both of which are challenging and time consuming tasks. A technical approach for alleviating the limitations of small datasets is to focus on learning a similarity metric on input data, rather than directly training a classifier. This includes \emph{embedding} the inputs; mapping the inputs to high dimensional real vectors, such that the easy to compute similarity between these vectors reflects the much more difficult to formalize similarity between the inputs \cite{abeysinghe_snake_2019, figueroa-mata_using_2020, koch_siamese_2015, hoffer_deep_2018, chopra_learning_2005, baldi_neural_1993}. 
One approach to learn this mapping is through the use of \emph{siamese networks} \cite{baldi_neural_1993}. Siamese networks are trained on pairs of inputs, with the goal of producing small distances between the embeddings of inputs from the same class and large distances between embeddings of the inputs from different classes. This approach was shown to work well with the comparatively small datasets used in biological applications, where the datasets often have many classes and few samples per class. For instance,  \citet{abeysinghe_snake_2019} utilize siamese networks to classify images of snakes, and \citet{figueroa-mata_using_2020} apply them to classify different plant species. 
\emph{Triplet networks} \cite{hoffer_deep_2018} build on this approach by training on three inputs: the anchor, a positive example from the same class as the anchor, and a negative example from a different class. 
The network is then trained to produce small distances between the embeddings of the anchor and the positive example and larger distances between the embeddings of the anchor and the negative example. 
This was shown to produce embeddings that capture the similarities in the underlying dataset in a way that is better suited for classification tasks \cite{hoffer_deep_2018, schroff_facenet_2015}. 
As triplet networks are learning the relative similarity of an input sample in relation to all other samples in the dataset, they were also shown to perform well with unbalanced datasets and in cases where multiple classes have only a small number of samples available \cite{koch_siamese_2015, schroff_facenet_2015}. 
We leverage this capability of triplet networks to learn meaningful representations to address the challenges posed by our small, imbalanced dataset, where for many classes only a small number of samples exists.

\subsection{Transfer Learning and Multi-Task Learning}
Training deep learning systems on small datasets can easily result in poor performance and overfitting. Two effective techniques to mitigate these issues are \emph{transfer learning} and \emph{multi-task learning}. 
In \emph{transfer learning}, also known as \emph{fine-tuning}, the network is first trained on a different, generally much larger dataset. The idea is that during this \emph{pre-training} phase, the network learns expressive intermediate features, which can be leveraged to achieve high performance on similar tasks \cite{oquab_learning_2014, pan_survey_2010}.
\emph{Multi-task learning} on the other hand, aims to enhance performance on a task by exploiting knowledge from different related tasks, for example by learning multiple tasks at the same time, while using a shared intermediate representation. This approach has been shown to improve generalization and model performance over training models for the tasks in isolation \cite{caruana_multitask_1997, thrun_is_1995, kendall_multi-task_2018}, and it can also help alleviate limitations from small datasets \cite{zhang_overview_2017}. A key challenge in this learning process is how to combine the multiple objectives into a single loss function. The naive approach of using a weighted sum requires careful tuning, as the different loss functions can have vastly different value ranges. To address this, \citet{kendall_multi-task_2018} propose a method to learn optimal loss weights during the training process.
In our system, we use a convolutional neural network that was pre-trained on the over 1 million images and 1000 classes of the ImageNet dataset \cite{deng_imagenet_2009} to extract expressive image features, helping us overcome the limitations of our small dataset, and providing a solid foundation for capturing meaningful image features from our data. Additionally, we observed that during training the loss values for embeddings and classification have very different ranges. Therefore, we adopt the approach proposed by \citet{kendall_multi-task_2018} to avoid the time-consuming search for optimal individual loss weights while simultaneously learning embeddings and classification.

\section{Dataset}
\begin{figure}[htb]
    \centering
    \includesvg[width=0.9\linewidth]{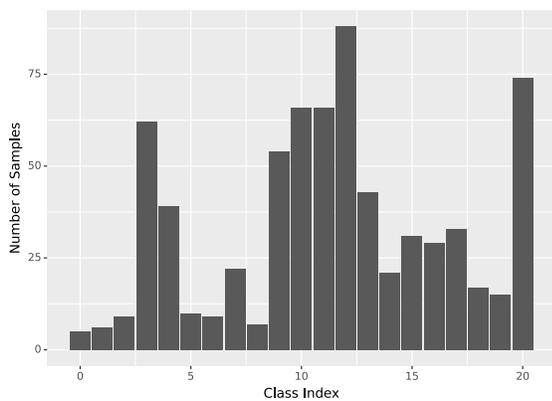}
    \caption{Samples per class in the dataset. The distribution is very imbalanced, most samples (88) are available for \emph{R. mostarensis}, least samples (5) are availabe for \emph{R. albanica}}
    \label{fig:class-counts}
\end{figure}

Our system is trained on and applied to a dataset consisting of 706 specimens of the genus \emph{Radomaniola}, distributed across 21 species, with between 5 and 88 specimens each. As illustrated by \cref{fig:class-counts}, the dataset is very imbalanced, with multiple classes represented by less than 10 samples.
For each specimen, we have a high-quality photograph available, as illustrated in \cref{fig:snails}, which we process by resizing to 224x224 pixels, mapping the values for each channel to values in the range from 0 to 1, and then normalizing the values so that across the complete dataset each channel has a mean of 0 and a standard deviation of 1.

In addition to the images, we have available multiple measurements of the specimen and the site where it was collected.
The specimen-specific measurements consist of shell length, shell width, aperture length, aperture width, width of body whorl, and width of penultimate whorl. 
Furthermore, for each specimen information on the site where it was collected, including country, geographic coordinates, habitat type in IUCN units \citepalias{iucn_international_union_for_conservation_of_nature_habitats_2012}, local temperature and precipitation obtained via coordinates from WorldClim database \cite{worldclimorg_worldclim_nodate}, geological information derived via coordinates from The Geological Map of Europe \cite{asch_igme_2005}, and elevation.

However, on a first analysis we found that this data was too specific.
In the dataset, there was exactly one species present at each collection site.
At the same time, very specific environmental measurements were available, such as latitude and longitude of the collection site, and temperature ranges at the collection site.
These measurements in turn allowed our system to shortcut the learning process by simply learning which measurements belong to which site and which species was collected at that site. 
It should, however, be noted that these are not necessarily confounding variables, as geographic location and environmental parameters, such as temperature, are important descriptors of species habitats.
In our case we found them simply measured too fine-grained to allow learning of useful features.
To avoid the system relying on shortcuts, we removed all features that allowed identification of a specific site, such as geographic coordinates, reducing the number of available measurements from 19 to 12 for each specimen. 

We validated the removal of features by training a random forest classifier on the remaining measurements with 80\% of data used for training and 20\% for testing. 
As the classifier was not able to reach above 90\% accuracy on the test set, we concluded that the reduced feature set did not allow for shortcuts anymore.
For comparison, training on various different subsets of the removed attributes allowed for perfect 100\% accurate classification. 
Investigating feature importances, we found that the country where the samples were collected proved to be a very strong predictor for the species. 
This also matches how the domain experts perform their classification process, as many species have specific countries where they are observed. 
Therefore, knowing that a sample comes from a certain country already allows the domain experts to exclude multiple species as very unlikely, and our ML system seems to follow a similar process.

Finally, we also have genetic data on the differences between species available.
We use data presented by \citet[Fig. 3A]{delicado_shell_2022}, where a clustering of species, acquired through Bayesian coalescence, is depicted.
In this clustering, the similarities are based on mutations in DNA, rather than morphology. 
It serves as a reference to understand which species should be more similar in shell morphology based on their evolutionary history.
We compiled this clustering into a pairwise distance matrix which indicates similarity between species in time units. 
The lower the distance, the more similarities the DNA of two species is sharing.

\section{System description}
\begin{figure*}[ht]
    \centering
    \includegraphics[width=0.95\linewidth]{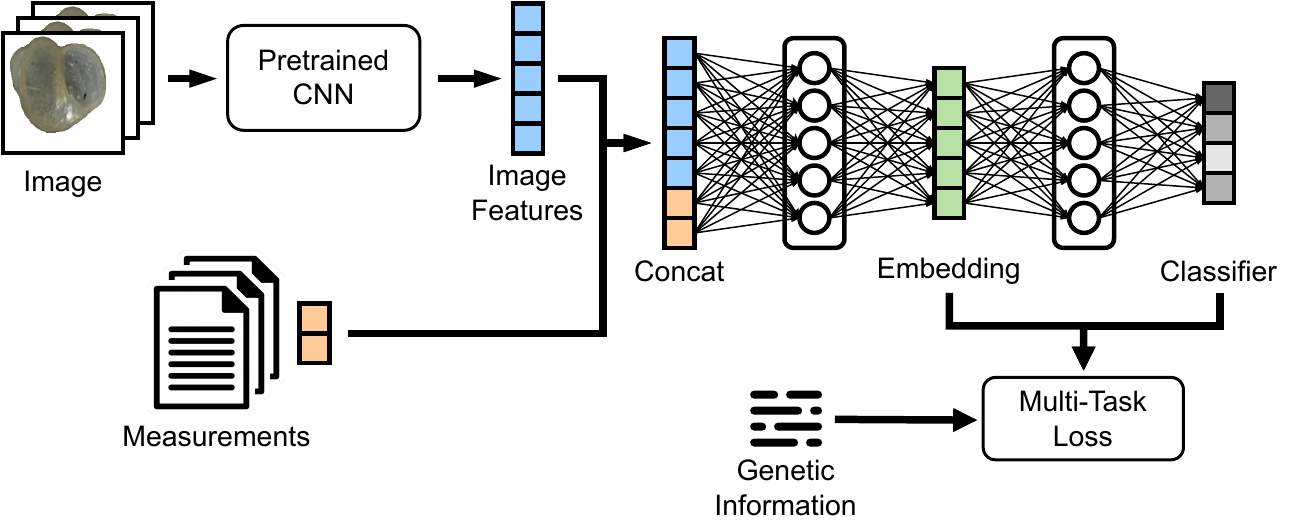}
    \caption{Architecture of the classification system. A pre-trained CNN extracts image features from the input images. The image features and the measurements are concatenated to create a joint representation. This joint representation is mapped to an embedding, which is then used to classify the input. Genetic information is only used during training, where it is used in the joint optimization of embeddings and classification results.}
    \label{fig:architecture}
\end{figure*}
As illustrated by \cref{fig:architecture}, our system consists of the following steps:
\begin{itemize}
    \item Image feature extraction,
    \item Integrating measurements into a joint representation,
    \item Learning embeddings from the joint representation,
    \item Using embeddings for classification.
\end{itemize}
We use a MobileNetV3-small network \cite{howard_searching_2019}, pre-trained on ImageNet \cite{deng_imagenet_2009}, to map input images to 32-dimensional image feature vectors. This comparatively small network allows for fast interference times and has low requirements for training and deployment. In preliminary experiments we also evaluated different ResNet \cite{he_deep_2015} and EfficientNet \cite{tan_efficientnetv2_2021} architectures, but the MobileNet architecture showed the best performance in our system.
The measurements are then integrated into a joint representation by concatenating the raw measurement values with the image feature vector, as illustrated by the \emph{Concat} layer in \cref{fig:architecture}.
This joint representation is then mapped to a 32-dimensional embedding, which is trained using the triplet loss \cite{hoffer_deep_2018}. For an input $x$, a positive example $x_+$ from the same class as $x$ and a negative example $x_-$ from a different class, the triplet loss is given as:
\begin{equation}
\label{eq:triplet}
    \mathcal{L}_{t}(x, x_+, x_-) = \mathit{ReLU}(d_{+} - d_{-} + \mathit{margin}) 
\end{equation}
Here, $d_{+} := \mathit{dist(x, x_+)}$ is the distance in embedding space between anchor and positive example, and $d_- := \mathit{dist(x, x_-)}$ the distance in embedding space between anchor and negative example. Optimizing this loss leads to embeddings where inputs from the same class are close together and inputs from different classes are at least $\mathit{margin}$ apart. In triplet networks, correct selection of triplets is essential for convergence \cite{schroff_facenet_2015}. Due to the limited size of our dataset, we can perform offline triplet selection, where after each epoch we compute good triplets of inputs over the whole dataset. For our triplets we use the hardest positive and a semi-hard negative example. This means that for each anchor input, we select the input whose embedding is the furthest away. We then select a random \emph{semi-hard} negative example so that the distances suffice the equation $d_- < d_+ + \mathit{margin}$, as this was shown to help mitigate model collapse \cite{schroff_facenet_2015}. While this leads to embeddings that clearly separate the different classes, we also aim for biologically meaningful embeddings that capture the similarities between species. To achieve this, we replaced the fixed margin in \cref{eq:triplet} with what we call a \emph{dynamic margin} $\mathit{margin(x_+, x_-)}$. The \emph{dynamic margin} depends on the classes of anchor / positive example and negative example and instead of a fixed value it returns a margin value that is proportional to the innate difference between the classes. In our system, we used genetic distance between species, informed by \citet[Fig. 3A]{delicado_shell_2022}, to capture these innate differences between classes, and we use cosine distance to measure distances in embedding space.

Finally, the embeddings are fed into a classification head, where the loss for input $x$ with true label $y$, $\mathcal{L}_c(x, y)$, is given by the standard cross-entropy loss for classification tasks.
We then train the network end-to-end to optimize both losses $\mathcal{L}_t$ and $\mathcal{L}_c$ simultaneously, with the approach described by \citet{kendall_multi-task_2018} to also learn appropriate weights for each loss. 
The final loss function for our system is given by the following equation, where both losses and their respective weights ($s_1, s_2$) are jointly optimized:
\begin{equation}
\begin{split}
    \mathcal{L}(x, x_+, x_-, y) =& \text{ }\mathrm{exp}(-s_1) \cdot \mathcal{L}_t(x, x_+, x_-) \\
    &+ \mathrm{exp}(-s_2) \cdot \mathcal{L}_c(x, y) \\
    &+ s_1 + s_2
\end{split}
\end{equation}
This combined loss helps us to leverage the ability of triplet networks and multi-task learning to learn meaningful representations from small sample sizes and imbalanced datasets. At the same time, the joint training ensures that the learnt representations are well suited for classification, and trains the classifier how to best use them for its task.

\section{Experiments}
We compare our system against multiple baselines to ensure that each level of added complexity genuinely improves classification performance. For this we evaluate the performance of the following configurations: (1) images, (2) images and measurements, (3) image and measurements, trained with triplet loss and classification loss, and (4) images and measurements, trained with dynamic margin triplet loss and classification loss. For each configuration we estimate performance with 5-fold stratified cross validation, with 80\% of samples used for training and 20\% for testing. We train with a batch size of 16 for 100 epochs using the Adam optimizer \cite{kingma_adam_2017} with a learning rate of 0.001 and a cosine annealing schedule \cite{loshchilov_sgdr_2017} that decays the learning to 0.00001 over the course of the 100 epochs. Given the varying class sizes, we balance the class distribution in the training set, but not in the test set to avoid introducing biases regarding class frequencies into the training of the classifier. Furthermore, we augment the training images in each batch by randomly adding Gaussian noise with a mean of 0 and standard deviation of 0.05 and randomly rotating each image an integer multiple of 90 degrees. We explicitly avoid the common technique of flipping images, as this would would transform the \emph{dextral} shell images into pseudo-sinistral shell images. Since such snails actually exist in nature, this image transformation would have serious biological implications, especially when applied to a dataset containing both dextral and sinistral shells in the future.

\begin{figure}[htb]
    \centering
    \includesvg[width=0.9\linewidth]{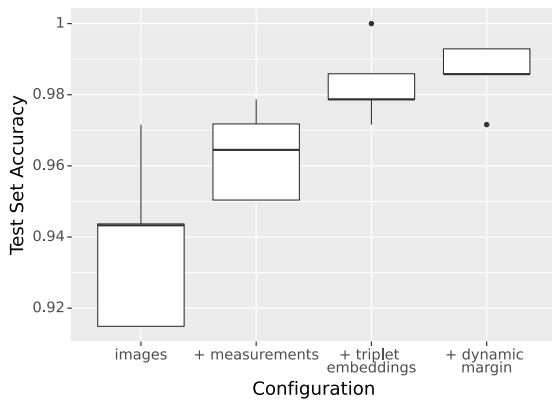}
    \caption{Test set classification accuracy for each of the configurations. Inclusion of additional modalities improves system performance, although the improvement from using genetic information with the dynamic margins is not statistically significant.}
    \label{fig:results}
\end{figure}

As seen in \cref{fig:results}, each additional level of complexity improves classification accuracy. Incorporating both images and measurements leads to better performance than using only images. Training the model on both triplet embeddings and classification leads to better results than training the model only on classification. Adding genetic species distances with the dynamic margin to the triplet embeddings also slightly improves the final classification accuracy over the standard triplet embeddings, though this improvement is minor and not statistically significant $(p>0.6)$. It is important to note that so far the genetic information is only used for optimizing the triplet loss during training and is not used during testing. However, with a final mean classification accuracy exceeding 98.5\%, the system is performing extraordinarily well and way above our initial expectations, considering the small dataset, unbalanced classes, and high visual similarity between classes. We also observe that, similar to the domain experts, our system can perform reasonably well with images alone, but including other modalities makes the classification process easier. 

\section{Conclusion and Outlook}
In this work we presented a system for the classification of \emph{Radomaniola} snails based on multimodal triplet networks and demonstrate that it can achieve expert-level classification accuracy. 
A big strength of our system is that it does not require thousands of images for success, but is able to achieve its high performance despite the small dataset with less than 1000 images, the very imbalanced class distributions where some classes have less than ten images available, and the very subtle differences between classes that are hardly visible to non-experts. 
In future work, we intend to incorporate different explanation techniques in the system. In close cooperation with the domain experts, we intend to use this to investigate what the system learns, if it can capture new information, whether the explanations are biologically sound, and which of the many different approaches to explain the decision process of deep neural networks is the most useful to them. 

Our classification system is designed to be a valuable tool for domain experts by reducing their workload, speeding up the classification process, and maintaining high accuracy. 
With interference times in the range of only a few seconds on current consumer hardware, even without GPU acceleration, it might even be possible to apply the system in the field in combination with a laptop and USB microscope.
Such a setup would enable preliminary classification directly at collection sites and would benefit ecologists and conservation planners who are mostly interested in identifying species in specific regions. 
Taxonomists or evolutionary biologists may still need to work in the lab for detailed anatomical or genetical analyses.
Given the difficulties that experts face in classifying \textit{Radomaniola} specimens, comprehensive documentation of shell and other anatomical features using a high-resolution binocular microscope in the laboratory is essential for meticulous species comparison. 
Here our ML system can significantly reduce the required effort by automating cross-species comparisons based on shell and geographic data.
We estimate that with the aid of the ML system, experts will be able to classify a specimen in minutes, compared to the hours it might take using traditional methods.

\bibliography{bibliography}

\begin{thebibliography}{31}
\providecommand{\natexlab}[1]{#1}
\providecommand{\url}[1]{\texttt{#1}}
\expandafter\ifx\csname urlstyle\endcsname\relax
  \providecommand{\doi}[1]{doi: #1}\else
  \providecommand{\doi}{doi: \begingroup \urlstyle{rm}\Url}\fi

\bibitem[Abeysinghe et~al.(2019)Abeysinghe, Welivita, and Perera]{abeysinghe_snake_2019}
Abeysinghe, C., Welivita, A., and Perera, I.
\newblock Snake {Image} {Classification} using {Siamese} {Networks}.
\newblock In \emph{Proceedings of the 3rd {International} {Conference} on {Graphics} and {Signal} {Processing}}, {ICGSP} '19, pp.\  8--12, New York, NY, USA, June 2019. Association for Computing Machinery.
\newblock ISBN 978-1-4503-7146-9.
\newblock \doi{10.1145/3338472.3338476}.
\newblock URL \url{https://dl.acm.org/doi/10.1145/3338472.3338476}.

\bibitem[Asch(2005)]{asch_igme_2005}
Asch, K.
\newblock {IGME} 5000: 1 : 5 {Million} {International} {Geological} {Map} of {Europe} and {Adjacent} {Areas}.
\newblock Technical report, BGR, Hannover, 2005.

\bibitem[Baldi \& Chauvin(1993)Baldi and Chauvin]{baldi_neural_1993}
Baldi, P. and Chauvin, Y.
\newblock Neural {Networks} for {Fingerprint} {Recognition}.
\newblock \emph{Neural Computation}, 5\penalty0 (3):\penalty0 402--418, May 1993.
\newblock ISSN 0899-7667, 1530-888X.
\newblock \doi{10.1162/neco.1993.5.3.402}.
\newblock URL \url{https://direct.mit.edu/neco/article/5/3/402-418/5704}.

\bibitem[Baltrušaitis et~al.(2017)Baltrušaitis, Ahuja, and Morency]{baltrusaitis_multimodal_2017}
Baltrušaitis, T., Ahuja, C., and Morency, L.-P.
\newblock Multimodal {Machine} {Learning}: {A} {Survey} and {Taxonomy}, August 2017.
\newblock URL \url{http://arxiv.org/abs/1705.09406}.
\newblock arXiv:1705.09406 [cs].

\bibitem[Boeters et~al.(2017)Boeters, Glöer, and Slavevska~Stamenković]{boeters_radomaniolagrossuana_2017}
Boeters, H.~D., Glöer, P., and Slavevska~Stamenković, V.
\newblock The {\emph{radomaniola}}/{\emph{grossuana}} group from the {Balkan} {Peninsula}, with a description of \emph{Grossuana maceradica} n. sp. and the designation of a neotype of \emph{Paludina hohenackeri} {Küster}, 1853 ({Caenogastropoda}: {Truncatelloidea}: {Hydrobiidae}).
\newblock \emph{Archiv für Molluskenkunde International Journal of Malacology}, 146\penalty0 (2):\penalty0 187--202, December 2017.
\newblock ISSN 1869-0963.
\newblock \doi{10.1127/arch.moll/146/187-202}.

\bibitem[Caruana(1997)]{caruana_multitask_1997}
Caruana, R.
\newblock Multitask {Learning}.
\newblock \emph{Machine Learning}, 28\penalty0 (1):\penalty0 41--75, July 1997.
\newblock ISSN 1573-0565.
\newblock \doi{10.1023/A:1007379606734}.
\newblock URL \url{https://doi.org/10.1023/A:1007379606734}.

\bibitem[Chopra et~al.(2005)Chopra, Hadsell, and LeCun]{chopra_learning_2005}
Chopra, S., Hadsell, R., and LeCun, Y.
\newblock Learning a {Similarity} {Metric} {Discriminatively}, with {Application} to {Face} {Verification}.
\newblock In \emph{2005 {IEEE} {Computer} {Society} {Conference} on {Computer} {Vision} and {Pattern} {Recognition} ({CVPR}'05)}, volume~1, pp.\  539--546, San Diego, CA, USA, 2005. IEEE.
\newblock ISBN 978-0-7695-2372-9.
\newblock \doi{10.1109/CVPR.2005.202}.
\newblock URL \url{http://ieeexplore.ieee.org/document/1467314/}.

\bibitem[Delicado \& Hauffe(2022)Delicado and Hauffe]{delicado_shell_2022}
Delicado, D. and Hauffe, T.
\newblock Shell features and anatomy of the springsnail genus \textit{{Radomaniola}} ({Caenogastropoda}: {Hydrobiidae}) show a different pace and mode of evolution over five million years.
\newblock \emph{Zoological Journal of the Linnean Society}, 196\penalty0 (1):\penalty0 393--441, August 2022.
\newblock ISSN 0024-4082, 1096-3642.
\newblock \doi{10.1093/zoolinnean/zlab121}.

\bibitem[Deng et~al.(2009)Deng, Dong, Socher, Li, Li, and Fei-Fei]{deng_imagenet_2009}
Deng, J., Dong, W., Socher, R., Li, L.-J., Li, K., and Fei-Fei, L.
\newblock {ImageNet}: {A} {Large}-{Scale} {Hierarchical} {Image} {Database}.
\newblock In \emph{{CVPR09}}, 2009.

\bibitem[Figueroa-Mata \& Mata-Montero(2020)Figueroa-Mata and Mata-Montero]{figueroa-mata_using_2020}
Figueroa-Mata, G. and Mata-Montero, E.
\newblock Using a {Convolutional} {Siamese} {Network} for {Image}-{Based} {Plant} {Species} {Identification} with {Small} {Datasets}.
\newblock \emph{Biomimetics}, 5\penalty0 (1):\penalty0 8, March 2020.
\newblock ISSN 2313-7673.
\newblock \doi{10.3390/biomimetics5010008}.
\newblock URL \url{https://www.mdpi.com/2313-7673/5/1/8}.

\bibitem[Gao et~al.(2020)Gao, Li, Chen, and Zhang]{gao_survey_2020}
Gao, J., Li, P., Chen, Z., and Zhang, J.
\newblock A {Survey} on {Deep} {Learning} for {Multimodal} {Data} {Fusion}.
\newblock \emph{Neural Computation}, 32\penalty0 (5):\penalty0 829--864, May 2020.
\newblock ISSN 0899-7667.
\newblock \doi{10.1162/neco_a_01273}.
\newblock URL \url{https://doi.org/10.1162/neco_a_01273}.

\bibitem[He et~al.(2015)He, Zhang, Ren, and Sun]{he_deep_2015}
He, K., Zhang, X., Ren, S., and Sun, J.
\newblock Deep {Residual} {Learning} for {Image} {Recognition}, December 2015.
\newblock URL \url{http://arxiv.org/abs/1512.03385}.
\newblock arXiv:1512.03385 [cs].

\bibitem[Hoffer \& Ailon(2018)Hoffer and Ailon]{hoffer_deep_2018}
Hoffer, E. and Ailon, N.
\newblock Deep metric learning using {Triplet} network, December 2018.
\newblock URL \url{http://arxiv.org/abs/1412.6622}.
\newblock arXiv:1412.6622 [cs, stat].

\bibitem[Hong et~al.(2015)Hong, Yu, Wan, Tao, and Wang]{hong_multimodal_2015}
Hong, C., Yu, J., Wan, J., Tao, D., and Wang, M.
\newblock Multimodal {Deep} {Autoencoder} for {Human} {Pose} {Recovery}.
\newblock \emph{IEEE Transactions on Image Processing}, 24\penalty0 (12):\penalty0 5659--5670, December 2015.
\newblock ISSN 1057-7149, 1941-0042.
\newblock \doi{10.1109/TIP.2015.2487860}.
\newblock URL \url{http://ieeexplore.ieee.org/document/7293666/}.

\bibitem[Howard et~al.(2019)Howard, Sandler, Chu, Chen, Chen, Tan, Wang, Zhu, Pang, Vasudevan, Le, and Adam]{howard_searching_2019}
Howard, A., Sandler, M., Chu, G., Chen, L.-C., Chen, B., Tan, M., Wang, W., Zhu, Y., Pang, R., Vasudevan, V., Le, Q.~V., and Adam, H.
\newblock Searching for {MobileNetV3}, 2019.
\newblock URL \url{https://arxiv.org/abs/1905.02244}.
\newblock Version Number: 5.

\bibitem[{(IUCN) International Union for Conservation of Nature}(2012)]{iucn_international_union_for_conservation_of_nature_habitats_2012}
{(IUCN) International Union for Conservation of Nature}.
\newblock Habitats {Classification} {Scheme} ({Version} 3.1), December 2012.
\newblock URL \url{https://www.iucnredlist.org/resources/habitat-classification-scheme}.

\bibitem[Kendall et~al.(2018)Kendall, Gal, and Cipolla]{kendall_multi-task_2018}
Kendall, A., Gal, Y., and Cipolla, R.
\newblock Multi-{Task} {Learning} {Using} {Uncertainty} to {Weigh} {Losses} for {Scene} {Geometry} and {Semantics}, April 2018.
\newblock URL \url{http://arxiv.org/abs/1705.07115}.
\newblock arXiv:1705.07115 [cs].

\bibitem[Kingma \& Ba(2017)Kingma and Ba]{kingma_adam_2017}
Kingma, D.~P. and Ba, J.
\newblock Adam: A method for stochastic optimization, 2017.

\bibitem[Koch et~al.(2015)Koch, Zemel, and Salakhutdinov]{koch_siamese_2015}
Koch, G., Zemel, R., and Salakhutdinov, R.
\newblock Siamese {Neural} {Networks} for {One}-shot {Image} {Recognition}.
\newblock \emph{ICML Deep Leaning Workshop}, 2015.

\bibitem[Loshchilov \& Hutter(2017)Loshchilov and Hutter]{loshchilov_sgdr_2017}
Loshchilov, I. and Hutter, F.
\newblock {SGDR}: {Stochastic} {Gradient} {Descent} with {Warm} {Restarts}, May 2017.
\newblock URL \url{http://arxiv.org/abs/1608.03983}.
\newblock arXiv:1608.03983 [cs, math].

\bibitem[Mroueh et~al.(2015)Mroueh, Marcheret, and Goel]{mroueh_deep_2015}
Mroueh, Y., Marcheret, E., and Goel, V.
\newblock Deep multimodal learning for {Audio}-{Visual} {Speech} {Recognition}.
\newblock In \emph{2015 {IEEE} {International} {Conference} on {Acoustics}, {Speech} and {Signal} {Processing} ({ICASSP})}, pp.\  2130--2134, South Brisbane, Queensland, Australia, April 2015. IEEE.
\newblock ISBN 978-1-4673-6997-8.
\newblock \doi{10.1109/ICASSP.2015.7178347}.
\newblock URL \url{http://ieeexplore.ieee.org/document/7178347/}.

\bibitem[Ngiam et~al.(2011)Ngiam, Khosla, Kim, Nam, Lee, and Ng]{ngiam_multimodal_2011}
Ngiam, J., Khosla, A., Kim, M., Nam, J., Lee, H., and Ng, A.~Y.
\newblock Multimodal deep learning.
\newblock In \emph{Proceedings of the 28th international conference on machine learning ({ICML}-11)}, pp.\  689--696, 2011.

\bibitem[Nguyen et~al.(2019)Nguyen, Kavuri, and Lee]{nguyen_multimodal_2019}
Nguyen, T.-L., Kavuri, S., and Lee, M.
\newblock A multimodal convolutional neuro-fuzzy network for emotion understanding of movie clips.
\newblock \emph{Neural Networks}, 118:\penalty0 208--219, October 2019.
\newblock ISSN 0893-6080.
\newblock \doi{10.1016/j.neunet.2019.06.010}.
\newblock URL \url{https://www.sciencedirect.com/science/article/pii/S0893608019301832}.

\bibitem[Oquab et~al.(2014)Oquab, Bottou, Laptev, and Sivic]{oquab_learning_2014}
Oquab, M., Bottou, L., Laptev, I., and Sivic, J.
\newblock Learning and {Transferring} {Mid}-level {Image} {Representations} {Using} {Convolutional} {Neural} {Networks}.
\newblock In \emph{2014 {IEEE} {Conference} on {Computer} {Vision} and {Pattern} {Recognition}}, pp.\  1717--1724, Columbus, OH, USA, June 2014. IEEE.
\newblock ISBN 978-1-4799-5118-5.
\newblock \doi{10.1109/CVPR.2014.222}.
\newblock URL \url{https://ieeexplore.ieee.org/document/6909618}.

\bibitem[Ouyang et~al.(2014)Ouyang, Chu, and Wang]{ouyang_multi-source_2014}
Ouyang, W., Chu, X., and Wang, X.
\newblock Multi-source {Deep} {Learning} for {Human} {Pose} {Estimation}.
\newblock pp.\  2329--2336, 2014.
\newblock URL \url{https://openaccess.thecvf.com/content_cvpr_2014/html/Ouyang_Multi-source_Deep_Learning_2014_CVPR_paper.html}.

\bibitem[Pan \& Yang(2010)Pan and Yang]{pan_survey_2010}
Pan, S.~J. and Yang, Q.
\newblock A {Survey} on {Transfer} {Learning}.
\newblock \emph{IEEE Transactions on Knowledge and Data Engineering}, 22\penalty0 (10):\penalty0 1345--1359, October 2010.
\newblock ISSN 1041-4347.
\newblock \doi{10.1109/TKDE.2009.191}.
\newblock URL \url{http://ieeexplore.ieee.org/document/5288526/}.

\bibitem[Schroff et~al.(2015)Schroff, Kalenichenko, and Philbin]{schroff_facenet_2015}
Schroff, F., Kalenichenko, D., and Philbin, J.
\newblock {FaceNet}: {A} {Unified} {Embedding} for {Face} {Recognition} and {Clustering}.
\newblock In \emph{2015 {IEEE} {Conference} on {Computer} {Vision} and {Pattern} {Recognition} ({CVPR})}, pp.\  815--823, June 2015.
\newblock \doi{10.1109/CVPR.2015.7298682}.
\newblock URL \url{http://arxiv.org/abs/1503.03832}.
\newblock arXiv:1503.03832 [cs].

\bibitem[Tan \& Le(2021)Tan and Le]{tan_efficientnetv2_2021}
Tan, M. and Le, Q.~V.
\newblock {EfficientNetV2}: {Smaller} {Models} and {Faster} {Training}, June 2021.
\newblock URL \url{http://arxiv.org/abs/2104.00298}.
\newblock arXiv:2104.00298 [cs].

\bibitem[Thrun(1995)]{thrun_is_1995}
Thrun, S.
\newblock Is {Learning} {The} n-th {Thing} {Any} {Easier} {Than} {Learning} {The} {First}?
\newblock In Touretzky, D., Mozer, M.~C., and Hasselmo, M. (eds.), \emph{Advances in {Neural} {Information} {Processing} {Systems}}, volume~8. MIT Press, 1995.
\newblock URL \url{https://proceedings.neurips.cc/paper_files/paper/1995/file/bdb106a0560c4e46ccc488ef010af787-Paper.pdf}.

\bibitem[worldclim.org()]{worldclimorg_worldclim_nodate}
worldclim.org.
\newblock {WorldClim} {Database}.
\newblock URL \url{https://www.worldclim.org/}.

\bibitem[Zhang \& Yang(2017)Zhang and Yang]{zhang_overview_2017}
Zhang, Y. and Yang, Q.
\newblock {An overview of multi-task learning}.
\newblock \emph{National Science Review}, 5\penalty0 (1):\penalty0 30--43, 09 2017.
\newblock ISSN 2095-5138.
\newblock \doi{10.1093/nsr/nwx105}.
\newblock URL \url{https://doi.org/10.1093/nsr/nwx105}.

\end{thebibliography}
\bibliographystyle{icml2024}

%%%%%%%%%%%%%%%%%%%%%%%%%%%%%%%%%%%%%%%%%%%%%%%%%%%%%%%%%%%%%%%%%%%%%%%%%%%%%%%
%%%%%%%%%%%%%%%%%%%%%%%%%%%%%%%%%%%%%%%%%%%%%%%%%%%%%%%%%%%%%%%%%%%%%%%%%%%%%%%
% APPENDIX
%%%%%%%%%%%%%%%%%%%%%%%%%%%%%%%%%%%%%%%%%%%%%%%%%%%%%%%%%%%%%%%%%%%%%%%%%%%%%%%
%%%%%%%%%%%%%%%%%%%%%%%%%%%%%%%%%%%%%%%%%%%%%%%%%%%%%%%%%%%%%%%%%%%%%%%%%%%%%%%
% \newpage
% \appendix
% \onecolumn

% You can have as much text here as you want. The main body must be at most $8$ pages long.
% For the final version, one more page can be added.
% If you want, you can use an appendix like this one.  

% The $\mathtt{\backslash onecolumn}$ command above can be kept in place if you prefer a one-column appendix, or can be removed if you prefer a two-column appendix.  Apart from this possible change, the style (font size, spacing, margins, page numbering, etc.) should be kept the same as the main body.
%%%%%%%%%%%%%%%%%%%%%%%%%%%%%%%%%%%%%%%%%%%%%%%%%%%%%%%%%%%%%%%%%%%%%%%%%%%%%%%
%%%%%%%%%%%%%%%%%%%%%%%%%%%%%%%%%%%%%%%%%%%%%%%%%%%%%%%%%%%%%%%%%%%%%%%%%%%%%%%

\end{document}